\documentclass[letterpaper, 10 pt, conference]{ieeeconf}  % Comment this line out if you need a4paper

\IEEEoverridecommandlockouts                              % This command is only needed if 
\usepackage{graphics} % for pdf, bitmapped graphics files
\usepackage{times} % assumes new font selection scheme installed
\usepackage{amsmath} % assumes amsmath package installed
\usepackage{amssymb}  % assumes amsmath package installed

\usepackage{caption}
\usepackage{graphicx}
\usepackage{multirow}
\usepackage{url}
\usepackage{hyperref}
\usepackage{cite}
\usepackage{subfig}
\usepackage{hyperref}

\usepackage{array}          % 用于表格列格式控制
\usepackage{booktabs}       % 用于专业表格线（\toprule, \midrule, \bottomrule等）
\usepackage{multirow}       % 用于单元格跨行合并（\multirow）
\usepackage{colortbl}       % 用于表格行/列颜色（\rowcolor, \cellcolor）
\usepackage{xcolor}         % 定义颜色（HTML颜色代码需要）
\usepackage{caption}        % 用于标题格式控制
\usepackage{graphicx}       % 用于\resizebox调整表格大小
\usepackage{makecell}       

\title{\LARGE \bf
TGM-VLA: Task-Guided Mixup for Sampling-Efficient and Robust Robotic Manipulation
}

\author{Fanqi Pu$^{1}$, Lei Jiang$^{2, *}$ and Wenming Yang$^{1, *}$
\thanks{$^{1}$Shenzhen International Graduate School, Tsinghua University, Shenzhen, China.
        Email: {\tt\small pfq23@mails.tsinghua.edu.cn} and {\tt\small yang.wenming@sz.tsinghua.edu.cn}}%
\thanks{$^{2}$The National and Local Co-Build Humanoid Robotics Innovation Center, Shanghai, China. 
        Email: {\tt\small jianglei@openloong.net}}%
\thanks{$*$ Corresponding Authors}
}

\begin{document}

\maketitle
\thispagestyle{empty}
\pagestyle{empty}

%%%%%%%%%%%%%%%%%%%%%%%%%%%%%%%%%%%%%%%%%%%%%%%%%%%%%%%%%%%%%%%%%%%%%%%%%%%%%%%%
\begin{abstract}

The performance of robotic imitation learning is fundamentally limited by data quality and training strategies. Prevalent sampling strategies on RLBench suffer from severe keyframe redundancy and imbalanced temporal distribution, leading to inefficient memory usage and unstable optimization. Moreover, reprojecting point clouds onto multi-view images with a black background—while more efficient than voxel-based methods—often causes dark objects to be indistinguishable and hard to manipulate. In this work, we propose a novel holistic framework that significantly improves both model performance and training efficiency. First, we redesign and optimize the keyframe sampling strategy, reducing memory consumption by 80\% and accelerating training speed by 5x. Second, we augment the model with a color inversion projection branch—a simple yet effective module that resolves the ambiguity of dark objects. Finally, we propose a task-guided mixup technique that dynamically fuses point clouds and action heatmaps according to task instructions, greatly improving robustness to distractors and performance in multi-goal scenarios. Extensive experiments demonstrate that our method achieves state-of-the-art performance with a 90.5\% success rate on RLBench and 68.8\% on the COLOSSEUM benchmark under challenging interference conditions. Our code and checkpoints are available at \url{https://github.com/PuFanqi23/TGM-VLA}.

\end{abstract}

%%%%%%%%%%%%%%%%%%%%%%%%%%%%%%%%%%%%%%%%%%%%%%%%%%%%%%%%%%%%%%%%%%%%%%%%%%%%%%%%
\section{INTRODUCTION}

Robotic imitation learning aims to enable robots to acquire complex manipulation skills through expert demonstration data. While conventional 2D image-based imitation methods \cite{zhao2023learning,chi2023diffusion,brohan2022rt,brohan2023rt, kimopenvla, chi2024diffusionpolicy, cheang2024gr, wu2023unleashing, li2025gr, shridhar2022cliport} struggle with perception and generalization in unstructured settings, recent advances in 3D Vision-Language-Action (VLA) Models have shown remarkable promise \cite{gervet2023act3d,shridhar2023perceiver,goyal2023rvt,goyal2024rvt,jia2024lift3d, yang2025fp3, li2025pointvla, qu2025spatialvla}. These models interpret language instructions to output actions from multi-view point clouds, demonstrating strong spatial generalization. Notably, frameworks like RVT \cite{goyal2023rvt} boost efficiency by reprojecting 3D point clouds into a 2D virtual images, reducing background interference.

Despite these advances, several critical limitations persist in current 3D VLA methods. First, conventional demonstration sampling strategies remain inefficient. Early sampling approaches like C2F-ARM\cite{james2022coarse} and PerACT \cite{shridhar2023perceiver} heavily oversample keyframes on RLBench benchmarks  \cite{james2020rlbench} to ensure the coverage of critic states and predict the next best action, a practice continued in later works such as RVT-2 \cite{goyal2024rvt} and SAM2ACT \cite{fang2025sam2act}. Although simple, this leads to substantial storage overhead and introduces severe temporal bias, with later frames disproportionately represented. Moreover, the choice of keyframes significantly affects performance; even large-scale models like BridgeVLA \cite{li2025bridgevla} can fail completely on certain tasks due to poorly chosen keyframes.

Second, the robustness of visual perception remains a bottleneck. For instance, while point cloud reprojection reduces computational cost, rendering points against a uniform background causes perceptual collapse when objects and background share similar colors, leading to failures even in training.  Furthermore, real-world scenes contain abundant task-irrelevant distractors, whereas current training typically includes only task-relevant objects. Consequently, these models underutilize textual instructions, becoming heavily dependent on visual inputs. During testing, the presence of novel distractors often distracts the model, leading to degraded accuracy or task failure.

\begin{figure}[t!]
  \centering
  \includegraphics[width=\columnwidth]{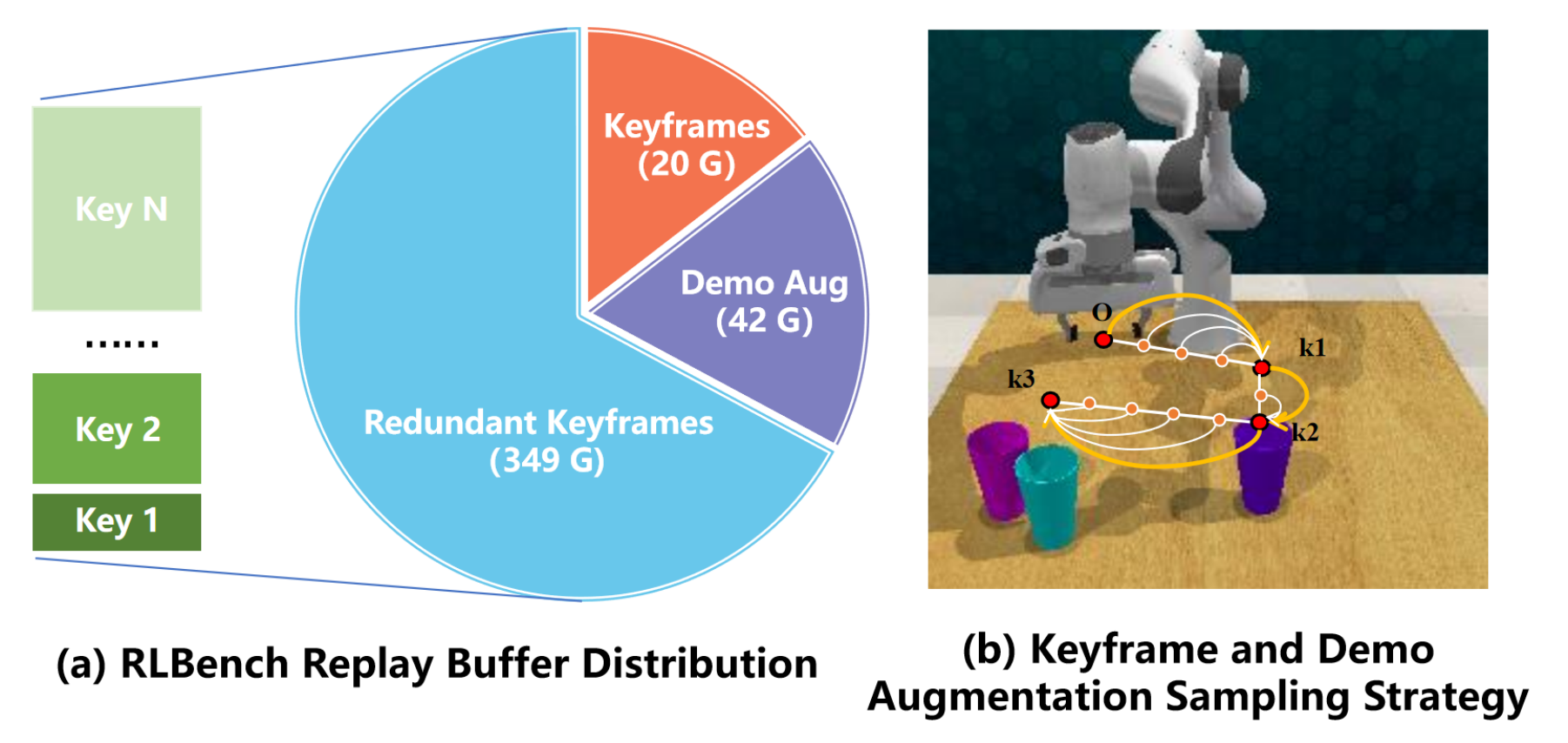}
    \caption{\textbf{Data Distribution and Sampling.} Red dots represent next target poses. Orange dots represent observed states sampled every 10 frames. Yellow arrows depict keyframe samples, while white arrows depict augmented samples. The original sampling strategy causes data redundancy and temporal imbalance, as each augmented sample forces repeated sampling of all subsequent keyframe samples.}
  \label{fig:sampling}
  \vspace{-15pt}
\end{figure}

\begin{figure*}[t!]  
  \centering
  \includegraphics[width=0.99\textwidth]{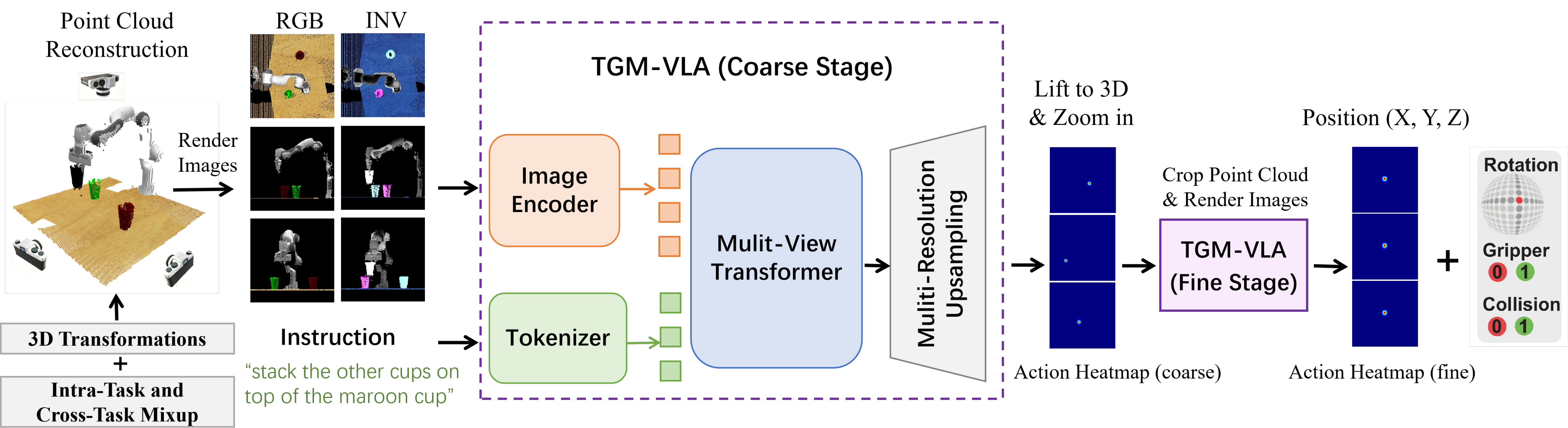}
  \caption{\textbf{Overview of TGM-VLA}. Given the current point cloud and a task instruction, TGM-VLA predicts the next key-frame pose. This model employs data augmentations including 3D transformations, intra-task, and cross-task mixup. At inference, it first renders point cloud into orthogonal views, enhanced with inverted-color images for improved contrast, to coarsely localize the area of interest. The second stage uses zoomed-in views to precisely predict the gripper pose.}
  \label{fig:model_architecture}
  \vspace{-8pt}
\end{figure*}

We address these challenges with following contributions:

\begin{itemize}
    \item An optimized keyframe sampling strategy reorganizes the dataset and uses cyclic alternate training, reducing replay buffer usage by over 80\% and accelerating training by 5× while improving performance.
    \item A color inversion point cloud projection branch is introduced to enhance contrast for dark objects, raising success rates by over 10\% in low-contrast scenes.
    \item A task-guided mixup strategy includes cross-task mixup to strengthen language-action alignment and intra-task mixup to capture multimodal action distributions.
    \item Our method achieves state-of-the-art results on two benchmarks: 90.5\% success on RLBench and 68.8\% on COLOSSEUM’s 14 interference settings.
\end{itemize}

\section{RELATED WORK}

\textbf{Learning robotic manipulation from demonstrations}. Imitation learning enables robots to acquire skills from expert demonstrations. Recent works have developed various approaches for different task settings, such as Diffusion Policy \cite{chi2023diffusion} for Push-T, ACT \cite{zhao2023learning} for bimanual manipulation, and RVT-2 \cite{goyal2024rvt} for language-conditioned tasks. In keyframe-based imitation learning, each datapoint in a demonstration is typically cast as a “predict the next (best) keyframe action” task \cite{james2022coarse, james2022q, liu2022auto}. Many approaches \cite{shridhar2023perceiver, johns2021coarse} employ keyframe discovery and demonstration augmentation to expand training data. However, these techniques often introduce noisy or sub-optimal samples, forcing heavy oversampling of keyframes to maintain performance. This results in highly redundant replay buffers, increased storage costs, and reduced training efficiency. Moreover, inappropriate keyframe selection can lead to critical failures such as motion repetition, inverse kinematics (IK) infeasibility, or obstacle avoidance breakdowns. Therefore, an optimized keyframe sampling strategy is essential to overcome these limitations.

\textbf{3D Vision-Language-Action (VLA) Models}. While 2D VLAs are efficient, they lack robustness in spatially complex tasks. These methods typically employ Transformers~\cite{vaswani2017attention} to process 2D visual inputs and directly predict 3D actions, often leveraging pre-trained vision-language models (VLMs) to build VLA models for complex manipulation learning~\cite{brohan2022rt, brohan2023rt, kimopenvla, black2410pi0, intelligence2025pi_, chi2024diffusionpolicy, li2023vision, li2025gr,cheang2024gr, zhao2023learning, wu2023unleashing}. Despite their effectiveness, such 2D image-based policies usually require extensive data collection, frequently demanding hundreds of trajectories per task to achieve reliable performance. This data inefficiency motivates the integration of 3D perception, leading to 3D VLA models that leverage structural scene understanding. Representative methods can be categorized into several groups. Point-cloud-based techniques, such as Act3D \cite{gervet2023act3d}, lift 2D features into 3D for action prediction. Voxel-based methods, including PerACT \cite{shridhar2023perceiver}, operate within a unified 3D voxel space. Hybrid designs aim to balance 3D capability with 2D efficiency; for example, RVT \cite{goyal2023rvt} and its successor RVT-2 \cite{goyal2024rvt} reproject 3D point clouds into 2.5D multi-view images, thereby avoiding direct 3D processing and accelerating training. SAM2ACT \cite{fang2025sam2act} and BridgeVLA \cite{li2025bridgevla} employ pre-trained backbones to predict heatmaps on 2D images projected from point clouds. However, this reprojection method struggles with low-contrast objects (e.g., dark items on black backgrounds). To mitigate this issue, we introduce a color inversion branch to strengthen perception, along with a task-guided point-cloud mixup strategy to improve language-action alignment and robustness against distractors.

\section{METHOD}

The proposed framework TGM-VLA learns a mapping \((\mathcal{P}, \mathcal{L}) \rightarrow \mathcal{A}\) from a 3D point cloud \(\mathcal{P}\) and a language instruction \(\mathcal{L}\) to an action command \(\mathcal{A}\). As shown in Fig.\ref{fig:model_architecture}, the multi-camera point cloud \(\mathbf{P} \in \mathbb{R}^{N \times 3}\) is first orthographically projected into multi-view images \(\{\mathbf{V}_i\}_{i=1}^K\), with inverted-color counterparts \(\{\overline{\mathbf{V}}_i\}\) generated to boost contrast. A visual encoder SAM2 \cite{ravi2024sam} extracts features from both \(\mathbf{V}_i\) and \(\overline{\mathbf{V}}_i\), while a text encoder CLIP \cite{radford2021learning} processes \(\mathcal{L}\). Per-view multimodal features \(\mathbf{F}_i = [\mathbf{F}_{\text{vis}}; \mathbf{F}_{\text{text}}]\) are fused via a multi-view Transformer \(\mathcal{T}\) using cross‑attention, yielding a global representation \(\mathbf{F}\).

An upsampling head \cite{teed2020raft} converts \(\mathbf{F}\) into orthogonal 2D action heatmaps \(\{\mathbf{H}_i\}\), whose back‑projection identifies the coarse 3D position with the highest score. This region is then cropped to obtain a refined, high‑resolution point cloud patch \(\mathbf{P}_{\text{refined}}\), and the same encoding‑prediction process is repeated. The model finally outputs an 8D action \(\mathcal{A}=(\mathbf{T}, s, c)\), which includes a 6‑DoF pose \(\mathbf{T} \in SE(3)\) (position \(\mathbf{t} \in \mathbb{R}^3\) and orientation \(\mathbf{R} \in SO(3)\)), a gripper state \(s\in\{0,1\}\), and a collision‑avoidance flag \(c\in\{0,1\}\), for subsequent motion planning.

\begin{figure*}[t!]
  \centering
  \includegraphics[width=\textwidth, trim={0 0 0 0}, clip]{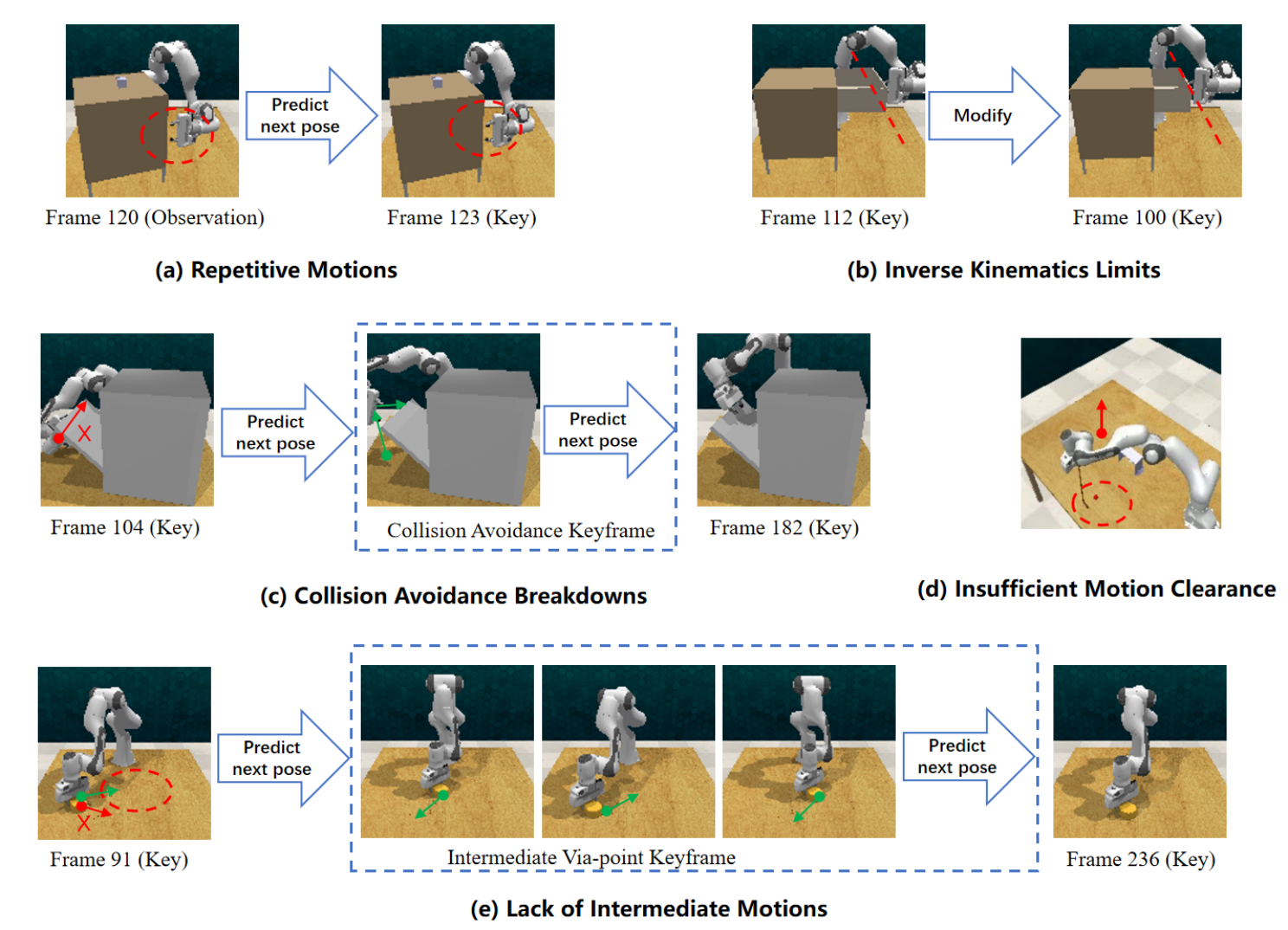}
    \caption{\textbf{Remedies for Failure Scenarios.} Each subfigure pairs a specific failure mode with our corrective measures.
    }
  \label{fig:redesign}
  \vspace{-8pt}
\end{figure*}

\subsection{Optimized Keyframe Sampling Strategy} 
In keyframe based robotic manipulation, the trajectory is represented as a sequence of key poses \( \mathcal{K} = \{ \mathbf{k}_1, \mathbf{k}_2, \dots, \mathbf{k}_N \} \). For instance, in a "stacking cups" task, keyframes are annotated at the pre-alignment and contact moments.  These poses are sampled from expert demonstration to construct a training replay buffer. Specifically, Current point cloud \(\mathbf{P}_t\) and language instruction \(\mathcal{L}\) are taken as input to predict the next keyframe pose \(\mathbf{k}_{t+1}\). 

The replay buffer consists of two sample types: \textbf{demo augmentation samples}, which use the point cloud at an observation frame as input and label it with the next keyframe pose to teach the model to progress from intermediate to target states; and \textbf{keyframe samples}, which take the point cloud at a keyframe as input and label it with the next keyframe pose to focus on transitions between critical states.

However, the conventional sampling strategy, which resamples all subsequent keyframes whenever an observation frame is selected, introduces two critical problems: \textbf{data redundancy}, as it generates duplicate samples that increase storage overhead and reduce dataset entropy; and \textbf{temporal imbalance}, since later keyframes are oversampled, causing the model to over‑emphasize later‑stage actions at the expense of early‑stage prediction.

To address these issues, we first removed redundant keyframes and employed a cyclic training strategy that alternates between keyframe samples and demo augmentation samples. However, after retraining, the model exhibited two unexpected failure modes:

\begin{itemize}
\item \textbf{Training Degradation}: For simple tasks (e.g., “pick and place”), the model initially reached 100\% success rate, but then declined substantially, stabilizing at a lower performance level ($60$--$70\%$).

\item \textbf{Training Deadlock}: For complex tasks (e.g., “empty dishwasher” or “wipe desk”), the model failed to learn from the start, maintaining a near-zero success rate. 
\end{itemize}

Analysis indicates that for imitation learning, data quality (“what to learn”) is as crucial as model architecture (“how to learn”). Thus, poor sampling becomes a bottleneck. These failures stem from two issues: the ineffectiveness of keyframes in capturing task states, and increased interference from erroneous samples under the new training strategy.

To determine whether a given sampling strategy is suitable for a task and to detect erroneous samples, we developed a stepwise diagnostic pipeline that monitors dynamic training success rates across task categories, enabling rapid identification of potential problem types. Based on success-rate patterns, training scenarios fall into three categories:

\begin{itemize}
\item \textbf{Scenario A}: Erroneous Samples --- Persistent decline in training success rate indicates erroneous samples in the replay buffer, requiring targeted data cleaning.

\item \textbf{Scenario B}: Ineffective Keyframes --- Steady overall success rate with certain instances at zero indicates ineffective keyframe selection, requiring respecification.

\item \textbf{Scenario C}: Generalization Limitations --- Stable training performance with a test-set gap confirms sampling effectiveness, requiring improved model generalization.
\end{itemize}

Based on preliminary diagnosis, we further record and analyze task failures such as repetitive motions and obstacle avoidance breakdowns. We observe that discarding redundant keyframes raised the sampling rate of demo augmentation frames, which contain erroneous samples. These previously diluted erroneous samples disrupted training and led to performance degradation or convergence stagnation.

Accordingly, we applied specific remedies: data cleaning for erroneous samples (Scenario A) and sampling redesign for ineffective keyframes (Scenario B). These measures eliminated systematic bias from prior sampling strategy, establishing a reliable data foundation for model training.

\textbf{Repetitive Motions} (e.g., "put item in drawer"). The original sampling strategy selects observation frames at fixed intervals \(\Delta t\) without imposing a minimum distance constraint between observation pose and next keyframe pose. These overly spatiotemporally close samples will induce the model to converge to a local optimum of near-static behaviors. To address this, we introduce a motion-saliency filter that retains only samples where the Euclidean distance exceeds a threshold. This adjustment improves training convergence and raises simple tasks from 60\% to a stable 100\%. 

\textbf{Inverse Kinematics Limits} (e.g., "open drawer"). When demonstration keyframes lie near workspace boundaries, even small perturbations during augmentation or inference can push target poses outside the feasible kinematic set, triggering IK failures. We mitigate this by retreating along the demonstrated trajectory by a fraction \(\alpha \in [0.8, 0.9]\) and creating a safety buffer before physical limits. Consequently, "open drawer" success increases from 88\% to 100\%.

\textbf{Collision Avoidance Breakdowns} (e.g., "empty dishwasher"). In cluttered or narrow environments, limited perception range and low-level motion planners often fail to prevent collisions. Therefore, we insert defensive via-point keyframes at high-risk zones, guiding the arm with a collision-free preparatory pose and decomposing risky motions into safer sub‑trajectories. This raises the success rate of "empty dishwasher" from 0\% to 36\%.

\textbf{Lack of Intermediate Motions} (e.g., "wipe desk"). Tasks with specific trajectory patterns such as zig-zag wiping suffer from weak path constraints when only current end-effector pose and next keyframe pose are provided. By inserting intermediate keyframes \(\{\mathbf{k}_{\text{via}}^{(i)}\}_{i=1}^{M-1}\) at curvature peaks of the desired trajectory, we enforce shape priors that compel the model to reproduce the intended motion trajectory. This raises the success rate of "wipe desk" from 0\% to 34\%.

\textbf{Insufficient Motion Clearance} (e.g., "hockey"). Some expert trajectories lack sufficient ground clearance at their lowest point. We present a defensive trajectory offset strategy that adopts a minimal 0.8 cm height offset to avoid the high-probability, high-severity ground-collision failure mode. This fine-grained adjustment raises the success rate of "hockey" from 20\% to over 64\%, underscoring the importance of micro-scale physical optimization.

Following data-level diagnosis, we executed data deduplication, anomaly cleansing, and sampling optimization. These steps improved overall task success, reduced replay buffer storage by 85\% (411 GB to 62 GB), and cut training time by 80\% (120 hours to 24 hours on 2 A6000 GPUs), significantly accelerating algorithm iteration.

\subsection{Color Inversion for Contrast Enhancement}

For 3D manipulation, methods like RVT \cite{goyal2023rvt} reconstruct point clouds and project them into several fixed orthogonal views. This avoids expensive 3D representations while providing multi-view perspectives beyond single images.

\begin{figure}[t!]
  \centering
  
  \includegraphics[width=\columnwidth]{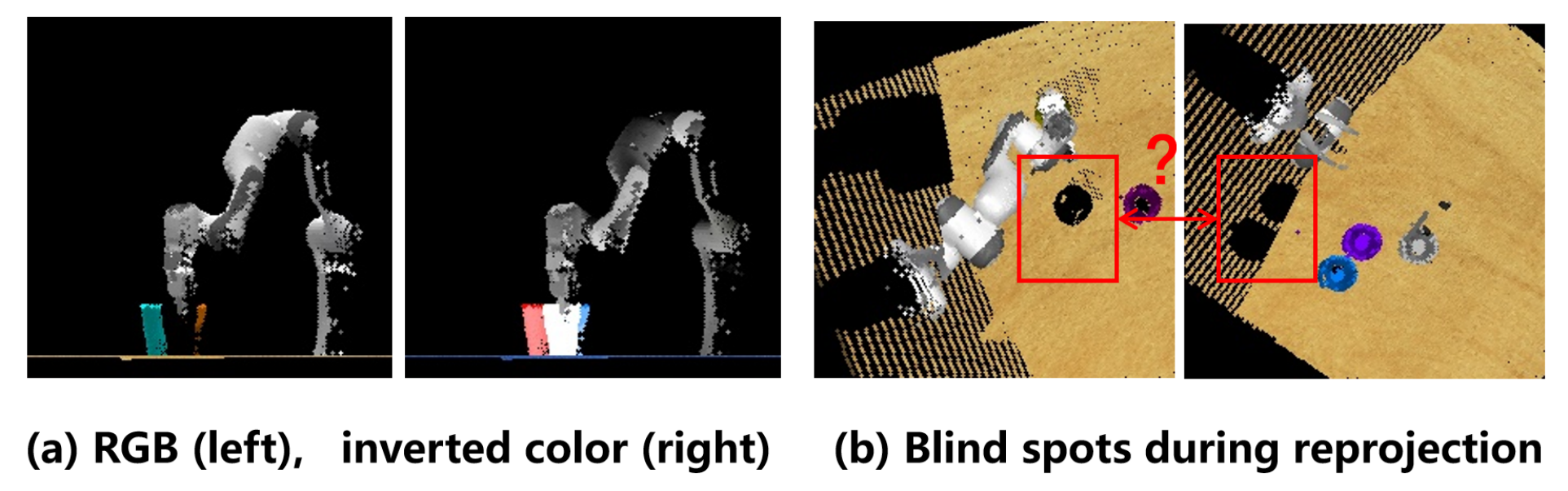}
  % \small
    \caption{{Low Contrast Problem and Color Inversion.} 
    }
  \label{fig:inversion}
  \vspace{-15pt}
\end{figure}

However, a key limitation persists: projections are typically rendered on a black background. As shown in Fig.\ref{fig:inversion} (a), dark objects such as dark cups blend into this background, losing contour and texture detail. This hinders detection and localization, so even if trained on such objects, the model fails to learn effective manipulation policies.

In complex scenes, the issue worsens. In Fig.\ref{fig:inversion} (b), dark, low-contrast regions can be mistaken for occlusion gaps or sensor blind spots, making it hard to tell empty areas from dark objects. 

To address this issue, we introduce \textbf{a color inversion branch} that enables the network to simultaneously process standard and inverted RGB streams. This ensures at least one stream exhibits high contrast against the black background, eliminating confusion from color similarity. Specifically, while generating standard orthogonal views, we create inverted views by replacing each pixel's color with \([255-R, 255-G, 255-B]\). Dark objects then appear bright, yielding clear contours. A feature fusion module integrates these complementary representations, preserving both true appearance and enhanced structure.

To further substantiate the necessity of color-space enhancement, we examine an alternative design that replaces the inversion branch with an auxiliary depth branch. Surprisingly, this variant yields no improvement in manipulation success on dark objects compared to the baseline without any augmentation. This finding reveals two critical aspects: first, the inversion branch is indispensable for resolving low-contrast failures; second, depth information, despite its geometric completeness, exhibits inherent limitations under the reprojection paradigm—it cannot compensate for the loss of color signals. We delve into these limitations from both data representation and network learning perspectives.

\begin{figure}
    \centering
    \includegraphics[width=0.99\linewidth]{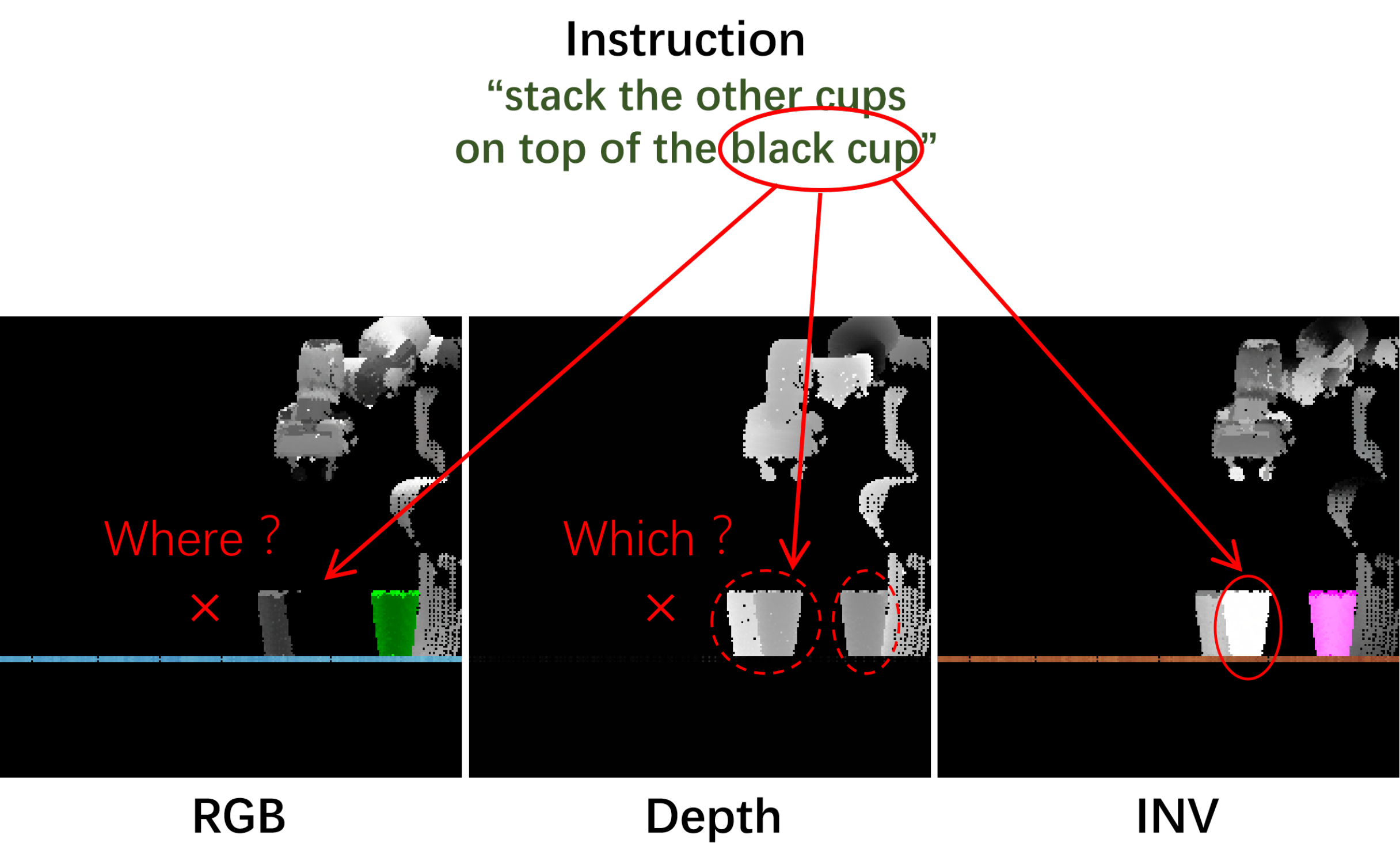}
    \caption{Alignment of Different Visual Modalities with Text Instructions.}
    \label{fig:depth_black}
    \vspace{-15pt}
\end{figure}

We attribute this ineffectiveness to two fundamental limitations of depth sensing within the reprojection paradigm. From a data perspective, depth maps capture geometry but lack color information, yet manipulation instructions often involve specific color semantics (e.g., "stack on top of the black cup"). As shown in Fig.~\ref{fig:depth_black}, even perfect 3D reconstruction provides no cue about object color, leaving the root cause—color-space signal degradation—unaddressed. From a network perspective, low contrast causes patch features of dark objects to approach zero vectors, rendering them indistinguishable from background in subsequent attention layers. Depth maps, residing in a different feature space, would require massive data-driven alignment with these degraded RGB features—a cross-modal association difficult to establish given limited dark-object examples. Thus, geometry alone cannot substitute for missing color signals; the problem fundamentally demands enhancing color contrast at the source.

In summary, our method generates both standard and inverted RGB views from point clouds, ensuring high-contrast representations across all colors against the black background. Geometry and color are orthogonal dimensions for describing objects—neither can substitute for the other. Therefore, the dark object perception problem can only be solved by enhancing the color signal itself. Our design effectively mitigates failures in low-contrast scenarios by preserving both true appearance and enhanced structure.

\begin{table*}[t]
\centering
\resizebox{\textwidth}{!}{ % Resize the table to fit within the text width
\begin{tabular}{|l|c|c|c|c|c|c|c|c|c|c|c|c}
\hline
\rowcolor[HTML]{D8E4FC} 
\multicolumn{3}{c}{\cellcolor[HTML]{D8E4FC}}                         & \multicolumn{2}{c}{\cellcolor[HTML]{D8E4FC}}                                                                                                                                                                       & \multicolumn{8}{c}{\cellcolor[HTML]{D8E4FC}}                                                                                                                                                                                                                                                                                                                                                                                                                                                                                                                                                                                                                                                                                                                                                                                                                  \\
\rowcolor[HTML]{D8E4FC} 
\multicolumn{3}{c}{\cellcolor[HTML]{D8E4FC}}                         & \multicolumn{2}{c}{\multirow{-2}{*}{\cellcolor[HTML]{D8E4FC}\textbf{Overall}}}                                                                                                                                               & \multicolumn{8}{c}{\multirow{-2}{*}{\cellcolor[HTML]{D8E4FC}\textbf{Task Success Rate (\%)}}}                                                                                                                                                                                                                                                                                                                                                                                                                                                                                                                                                                                                                                                                                                                                           \\
\cmidrule(lr){4-5} \cmidrule(l){6-13} 
\rowcolor[HTML]{D8E4FC} 
\multicolumn{3}{c}{\cellcolor[HTML]{D8E4FC}}                         & \cellcolor[HTML]{D8E4FC}                                                                                   & \cellcolor[HTML]{D8E4FC}                                                                               & \cellcolor[HTML]{D8E4FC}                                                                             & \cellcolor[HTML]{D8E4FC}                                                                            & \cellcolor[HTML]{D8E4FC}                                                                             & \cellcolor[HTML]{D8E4FC}                                                                                & \cellcolor[HTML]{D8E4FC}                                                                              & \cellcolor[HTML]{D8E4FC}                                                                            & \cellcolor[HTML]{D8E4FC}                                                                                  & \cellcolor[HTML]{D8E4FC}                                                                              \\
\rowcolor[HTML]{D8E4FC} 
\multicolumn{3}{c}{\multirow{-2}{*}{\cellcolor[HTML]{D8E4FC}\textbf{Models}}} & \multirow{-2}{*}{\cellcolor[HTML]{D8E4FC}\textbf{\begin{tabular}[c]{@{}c@{}}Avg.\\SR (\%) $\uparrow$ \end{tabular}}} & \multirow{-2}{*}{\cellcolor[HTML]{D8E4FC}\textbf{\begin{tabular}[c]{@{}c@{}}Reported\\Hardware \end{tabular}}}   & \multirow{-2}{*}{\cellcolor[HTML]{D8E4FC}\begin{tabular}[c]{@{}c@{}}Close\\Jar\end{tabular}}   & \multirow{-2}{*}{\cellcolor[HTML]{D8E4FC}\begin{tabular}[c]{@{}c@{}}Drag\\Stick\end{tabular}} & \multirow{-2}{*}{\cellcolor[HTML]{D8E4FC}\begin{tabular}[c]{@{}c@{}}Insert\\Peg\end{tabular}}  & \multirow{-2}{*}{\cellcolor[HTML]{D8E4FC}\begin{tabular}[c]{@{}c@{}}Meat off\\Grill\end{tabular}} & \multirow{-2}{*}{\cellcolor[HTML]{D8E4FC}\begin{tabular}[c]{@{}c@{}}Open\\Drawer\end{tabular}}  & \multirow{-2}{*}{\cellcolor[HTML]{D8E4FC}\begin{tabular}[c]{@{}c@{}}Place\\Cups\end{tabular}} & \multirow{-2}{*}{\cellcolor[HTML]{D8E4FC}\begin{tabular}[c]{@{}c@{}}Place\\Wine\end{tabular}}       & \multirow{-2}{*}{\cellcolor[HTML]{D8E4FC}\begin{tabular}[c]{@{}c@{}}Push\\Buttons\end{tabular}} \\

\midrule
\multicolumn{3}{c}{PerAct~\cite{shridhar2023perceiver}}                   & 49.4                                                                                                       & 8V100$\times$16d                                                                    & 55.2$\pm$4.7                                                                                             & 89.6$\pm$4.1                                                                                            & 5.6$\pm$4.1                                                                                              & 70.4$\pm$2.0                                                                                                & 88.0$\pm$5.7                                                                                              & 2.4$\pm$3.2                                                                                             & 44.8$\pm$7.8                                                                                                  & 92.8$\pm$3.0                                                                                              \\
\rowcolor[HTML]{E7E6E6} 
\multicolumn{3}{c}{\cellcolor[HTML]{E7E6E6}RVT~\cite{goyal2023rvt}}                      & 62.9                                                                                                       & 8V100$\times$1d                                                                     & 52.0$\pm$2.5                                                                                             & 99.2$\pm$1.6                                                                                            & 11.2$\pm$3.0                                                                                             & 88.0$\pm$2.5                                                                                                & 71.2$\pm$6.9                                                                                              & 4.0$\pm$2.5                                                                                             & 91.0$\pm$5.2                                                                                                  & \textbf{100.0$\pm$0.0}                                                                                    \\
\multicolumn{3}{c}{RVT-2~\cite{goyal2024rvt}}                    & 81.4                                                                                                       & 8V100$\times$20h                                                                  & \textbf{100.0$\pm$0.0}                                                                                   & 99.0$\pm$1.7                                                                                            & 40.0$\pm$0.0                                                                                             & 99.0$\pm$1.7                                                                                                & 74.0$\pm$11.8                                                                                             & 38.0$\pm$4.5                                                                                            & \textbf{95.0$\pm$3.3}                                                                                         & \textbf{100.0$\pm$0.0}                                                                                    \\
\rowcolor[HTML]{E7E6E6} 
\multicolumn{3}{c}{\cellcolor[HTML]{E7E6E6}SAM2ACT~\cite{fang2025sam2act}}                        & 86.8                                                                                                       & \begin{tabular}[c]{@{}c@{}}8H100$\times$12h or\\2A6000$\times$5d\end{tabular}                                                                 & 99.0$\pm$2.0                                                                                             & 99.0$\pm$2.0                                                                                            & 84.0$\pm$5.7                                                                                             & 98.0$\pm$2.3                                                                                                & 83.0$\pm$6.0                                                                                              & 47.0$\pm$6.0                                                                                            & 93.0$\pm$3.8                                                                                                  & \textbf{100.0$\pm$0.0}                                                                                    \\
\multicolumn{3}{c}{BridgeVLA~\cite{li2025bridgevla}}                                  & 88.2                                                                                              & 48H100$\times$20h                                                                   & \textbf{100.0$\pm$0.0}                                                                                   & \textbf{100.0$\pm$0.0}                                                                                  & 88.0$\pm$2.8                                                                                             & \textbf{100.0$\pm$0.0}                                                                                      & \textbf{100.0$\pm$0.0}                                                                                    & 58.4$\pm$10.0                                                                                  & 88.0$\pm$2.8                                                                                                  & 98.4$\pm$2.2                                                                                              \\
\rowcolor[HTML]{E7E6E6} 
\multicolumn{3}{c}{\cellcolor[HTML]{E7E6E6}\textbf{TGM-VLA w/o CTM}}                    & 88.2                                                                                                       & -                                                                   & 99.0$\pm$2.0                                                                                             & 98.0$\pm$2.3                                                                                            & 92.0$\pm$3.3                                                                                             & 96.0$\pm$3.3                                                                                                & \textbf{100.0$\pm$0.0}                                                                                    & 47.0$\pm$6.9                                                                                            & 94.0$\pm$2.0                                                                                                  & \textbf{100.0$\pm$0.0}                                                                                    \\
\multicolumn{3}{c}{\textbf{TGM-VLA w/o ITM}}                                  & 88.8                                                                                              & -                                                                   & 98.0$\pm$2.3                                                                                             & \textbf{100.0$\pm$0.0}                                                                                  & 91.0$\pm$3.8                                                                                             & 99.0$\pm$2.0                                                                                                & \textbf{100.0$\pm$0.0}                                                                                    & 50.0$\pm$4.0                                                                                            & 92.0$\pm$3.3                                                                                                  & \textbf{100.0$\pm$0.0}                                                                                              \\
\rowcolor[HTML]{E7E6E6} 
\multicolumn{3}{c}{\cellcolor[HTML]{E7E6E6}\textbf{TGM-VLA }}                    & \textbf{90.5}                                                                                                       & \textbf{2A6000$\times$1d}                                                                   & 99.0$\pm$2.0                                                                                             & \textbf{100.0$\pm$0.0}                                                                                  & \textbf{94.0$\pm$2.3}                                                                                    & 98.0$\pm$2.3                                                                                                & \textbf{100.0$\pm$0.0}                                                                                    & \textbf{60.0$\pm$7.3}                                                                                  & 94.0$\pm$2.3                                                                                                  & \textbf{100.0$\pm$0.0}                                                                                    \\
\bottomrule
\rowcolor[HTML]{D8E4FC} 
\multicolumn{3}{c}{\cellcolor[HTML]{D8E4FC}}                         & \cellcolor[HTML]{D8E4FC}                                                                                   & \cellcolor[HTML]{D8E4FC}                                                                               & \cellcolor[HTML]{D8E4FC}                                                                             & \cellcolor[HTML]{D8E4FC}                                                                            & \cellcolor[HTML]{D8E4FC}                                                                             & \cellcolor[HTML]{D8E4FC}                                                                                & \cellcolor[HTML]{D8E4FC}                                                                              & \cellcolor[HTML]{D8E4FC}                                                                            & \cellcolor[HTML]{D8E4FC}                                                                                  & \cellcolor[HTML]{D8E4FC}                                                                              \\
\rowcolor[HTML]{D8E4FC} 
\multicolumn{3}{c}{\multirow{-2}{*}{\cellcolor[HTML]{D8E4FC}\textbf{Models}}}       & \multirow{-2}{*}{\cellcolor[HTML]{D8E4FC}\begin{tabular}[c]{@{}c@{}}Put in\\Cupboard\end{tabular}}   & \multirow{-2}{*}{\cellcolor[HTML]{D8E4FC}\begin{tabular}[c]{@{}c@{}}Put in\\Drawer\end{tabular}} & \multirow{-2}{*}{\cellcolor[HTML]{D8E4FC}\begin{tabular}[c]{@{}c@{}}Put in\\Safe\end{tabular}} & \multirow{-2}{*}{\cellcolor[HTML]{D8E4FC}\begin{tabular}[c]{@{}c@{}}Screw\\Bulb\end{tabular}} & \multirow{-2}{*}{\cellcolor[HTML]{D8E4FC}\begin{tabular}[c]{@{}c@{}}Slide\\Block\end{tabular}} & \multirow{-2}{*}{\cellcolor[HTML]{D8E4FC}\begin{tabular}[c]{@{}c@{}}Sort\\Shape\end{tabular}}     & \multirow{-2}{*}{\cellcolor[HTML]{D8E4FC}\begin{tabular}[c]{@{}c@{}}Stack\\Blocks\end{tabular}} & \multirow{-2}{*}{\cellcolor[HTML]{D8E4FC}\begin{tabular}[c]{@{}c@{}}Stack\\Cups\end{tabular}} & \multirow{-2}{*}{\cellcolor[HTML]{D8E4FC}\begin{tabular}[c]{@{}c@{}}Sweep to\\Dustpan\end{tabular}} & \multirow{-2}{*}{\cellcolor[HTML]{D8E4FC}\begin{tabular}[c]{@{}c@{}}Turn\\Tap\end{tabular}}     \\
\midrule
\multicolumn{3}{c}{PerAct~\cite{shridhar2023perceiver}}                   & 28.0$\pm$4.4                                                                                                   & 51.2$\pm$4.7                                                                                               & 84.0$\pm$3.6                                                                                             & 17.6$\pm$2.0                                                                                            & 74.0$\pm$13.0                                                                                            & 16.8$\pm$4.7                                                                                                & 26.4$\pm$3.2                                                                                              & 2.4$\pm$2.0                                                                                             & 52.0$\pm$0.0                                                                                                  & 88.0$\pm$4.4                                                                                              \\
\rowcolor[HTML]{E7E6E6} 
\multicolumn{3}{c}{\cellcolor[HTML]{E7E6E6}RVT~\cite{goyal2023rvt}}                      & 49.6$\pm$3.2                                                                                                   & 88.0$\pm$5.7                                                                                               & 91.2$\pm$3.0                                                                                             & 48.0$\pm$5.7                                                                                            & 81.6$\pm$5.4                                                                                             & 36.0$\pm$2.5                                                                                                & 28.8$\pm$3.9                                                                                              & 26.4$\pm$8.2                                                                                            & 72.0$\pm$0.0                                                                                                  & 93.6$\pm$4.1                                                                                              \\
\multicolumn{3}{c}{RVT-2~\cite{goyal2024rvt}}                    & 66.0$\pm$4.5                                                                                                   & 96.0$\pm$0.0                                                                                               & 96.0$\pm$2.8                                                                                             & 88.0$\pm$4.9                                                                                   & 92.0$\pm$2.8                                                                                             & 35.0$\pm$7.1                                                                                                & \textbf{80.0$\pm$2.8}                                                                                     & 69.0$\pm$5.9                                                                                            & \textbf{100.0$\pm$0.0}                                                                                        & \textbf{99.0$\pm$1.7}                                                                                              \\
\rowcolor[HTML]{E7E6E6} 
\multicolumn{3}{c}{\cellcolor[HTML]{E7E6E6}SAM2ACT~\cite{fang2025sam2act}}                        & 75.0$\pm$3.8                                                                                                   & 99.0$\pm$2.0                                                                                               & 98.0$\pm$2.3                                                                                             & 89.0$\pm$2.0                                                                                            & 86.0$\pm$4.0                                                                                             & 64.0$\pm$4.6                                                                                                & 76.0$\pm$8.6                                                                                              & 78.0$\pm$4.0                                                                                            & 99.0$\pm$2.0                                                                                                  & 96.0$\pm$5.7                                                                                              \\
\multicolumn{3}{c}{BridgeVLA~\cite{li2025bridgevla}}                                  & 73.6$\pm$4.6                                                                                                   & 99.2$\pm$1.8                                                                                               & \textbf{99.2$\pm$1.8}                                                                                             & 87.2$\pm$6.6                                                                                            & 96.0$\pm$2.8                                                                                             & 60.8$\pm$7.7                                                                                                & 76.8$\pm$8.7                                                                                              & 81.6$\pm$3.6                                                                                            & 87.2$\pm$1.8                                                                                                  & 92.8$\pm$3.3                                                                                             \\
\rowcolor[HTML]{E7E6E6} 
\multicolumn{3}{c}{\cellcolor[HTML]{E7E6E6}\textbf{TGM-VLA w/o CTM}}                    & 80.0$\pm$3.3                                                                                                   & 95.0$\pm$2.0                                                                                               & 94.0$\pm$2.3                                                                                             & 88.0$\pm$4.6                                                                                            & 97.0$\pm$2.0                                                                                             & 73.0$\pm$3.8                                                                                                & 62.0$\pm$2.3                                                                                              & 79.0$\pm$8.2                                                                                            & 98.0$\pm$2.3                                                                                                  & 96.0$\pm$4.6                                                                                             \\
\multicolumn{3}{c}{\textbf{TGM-VLA w/o ITM}}                                  & 82.0$\pm$5.2                                                                                                   & 98.0$\pm$2.3                                                                                               & 95.0$\pm$3.8                                                                                             & 86.0$\pm$6.0                                                                                            & \textbf{99.0$\pm$2.0}                                                                                             & 78.0$\pm$7.3                                                                                                & 65.0$\pm$3.8                                                                                              & 82.0$\pm$6.5                                                                                            & 92.0$\pm$3.3                                                                                                  & 96.0$\pm$2.3                                                                                              \\
\rowcolor[HTML]{E7E6E6} 
\multicolumn{3}{c}{\cellcolor[HTML]{E7E6E6}\textbf{TGM-VLA }}                    & \textbf{83.0$\pm$2.0}                                                                                               & \textbf{100.0$\pm$0.0}                                                                                      & 96.0$\pm$4.6                                                                                    & \textbf{89.0$\pm$3.8}                                                                                            & \textbf{99.0$\pm$2.0}                                                                                   & \textbf{80.0$\pm$5.7}                                                                                       & 70.0$\pm$4.0                                                                                              & \textbf{87.0$\pm$2.0}                                                                                   & 88.0$\pm$4.6                                                                                                  & 93.0$\pm$3.8                                                                                             \\
\bottomrule
\end{tabular}
}
\vspace{3pt}
\caption{\textbf{Results on RLBench.} 
This table compares our TGM-VLA with prior works and ablations. “Avg. SR” is the average success rate over 18 tasks (higher is better). “Reported Hardware” lists the training resources (GPUs$\times$Type$\times$Time). “w/o CTM” and “w/o ITM” denote ablations without Cross-Task and Intra-Task Mixup, respectively. }
\label{tab:rlbench_tab}
\vspace{-15pt} 
\end{table*}

\subsection{Task-Guided Mixup for Input-Output Alignment}

In real-world robotic operations, irrelevant objects often interfere with the model’s perception of key regions and trigger erroneous actions. This stems from training on simplified scenes with only task-relevant objects, which makes the model over-rely on visual features and underutilize language guidance—even without language inputs, it still can achieve high success rates in many tasks. Furthermore, in tasks involving multiple targets (such as stacking cups),  the model predicts actions for a single target at a time and treats others as negative samples, confining it to a fixed operation sequence. Consequently, demonstrations with varying sequences create conflicting samples that impede model convergence.

We aim to enhance scene diversity without collecting new data, but traditional vision-language-action models operate in a continuous action space defined by joint angles or end-effector poses, which cannot output actions targeting multiple distinct locations simultaneously. Directly mixing samples from different tasks can easily lead to confusion in action representations, while simply averaging multiple target positions may yield physically implausible actions.

In contrast, methods such as RVT \cite{goyal2023rvt} reformulate action prediction as multi-view 2D probability heatmap prediction, allowing alignment between the image space and the action heatmap space. Under this paradigm, action outputs possess spatial additivity: multiple target positions can be integrated into a multi-peak heatmap, enabling the model to learn multiple feasible solutions simultaneously.

Therefore, we propose a task-guided point cloud mixup strategy, which is divided into intra-task mixup and cross-task mixup based on task types and textual instructions.

\textbf{Intra-task mixup} addresses tasks where a single instruction \(\mathcal{L}\) corresponds to multiple valid actions. During training, we blend point clouds from samples sharing the same instruction while directly concatenating their action heatmaps. Formally, given two samples \((\mathbf{P}_1 \in \mathbb{R}^{N_1 \times 3}, \mathcal{L}, \mathbf{H}_1)\) and \((\mathbf{P}_2 \in \mathbb{R}^{N_2 \times 3}, \mathcal{L}, \mathbf{H}_2)\), the augmented sample is:

\begin{equation}
\tilde{\mathbf{P}} = \mathrm{Concat}(\mathbf{P}_1, \mathbf{P}_2) \in \mathbb{R}^{(N_1+N_2) \times 3}, \ 
\tilde{\mathbf{H}} =\mathbf{H}_1 + \mathbf{H}_2
\end{equation}

Unlike regression-based methods that converge to a single optimal action, this strategy provides multimodal supervision. Consequently, the model learns to predict multi-peak heatmap distributions, capturing a set of plausible action locations \(\{\mathbf{p}_k\}\) rather than a single point \(\mathbf{p}^*\). This improves the model's ability to differentiate between category-level semantic abstraction and instance-specific actions. As a result, it enhances adaptability in multi-target decision-making.

\begin{figure}[]
  \centering
  \includegraphics[width=\columnwidth]{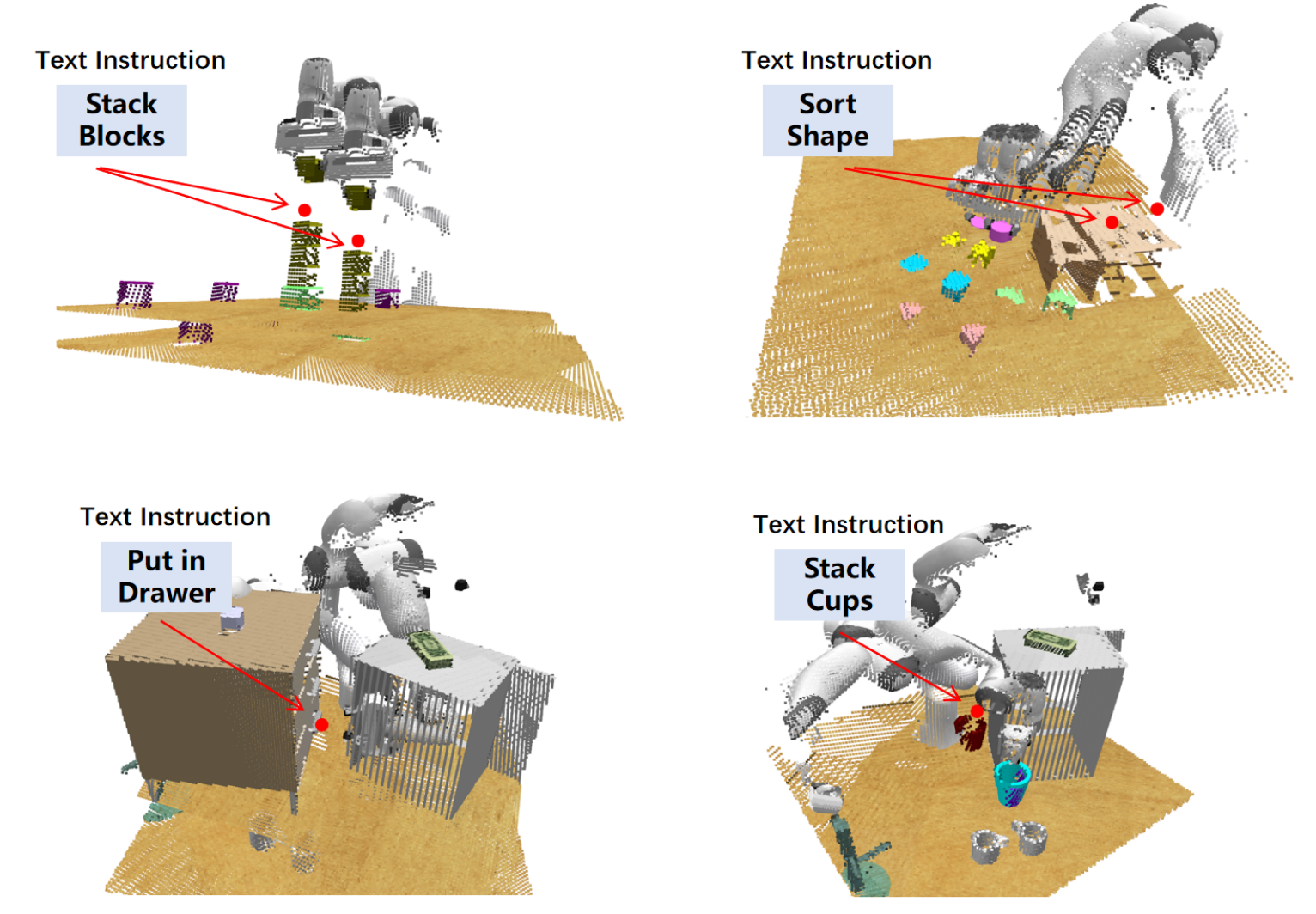}
    \caption{ Illustration of Intra-Task Mixup (top row) and Cross-Task Mixup (bottom row).
    }
  \label{fig:tgm}
  \vspace{-15pt}
\end{figure}

\textbf{Cross-task mixup} improves instruction following and attentional focus in cluttered scenes. It concatenates point clouds from \(M\) samples with distinct task instructions while retaining only the heatmap corresponding to the current instruction. Formally, given a set of samples \(\{(\mathbf{P}_m \in \mathbb{R}^{N_m \times 3}, \mathcal{L}_m, \mathbf{H}_m)\}_{m=1}^M\) with \(\mathcal{L}_1 \neq \mathcal{L}_m\) for \(m \neq 1\), the augmented sample is constructed as:

\begin{equation}
\tilde{\mathbf{P}} = \mathrm{Concat}(\mathbf{P}_1, \mathbf{P}_2, \dots, \mathbf{P}_M), \\\
\tilde{\mathcal{L}} = \mathcal{L}_1, \\\
\tilde{\mathbf{H}} = \mathbf{H}_1
\end{equation}

This simulates environments with task-irrelevant objects and enforces label preservation: the model must rely solely on \(\mathcal{L}_1\) to identify relevant regions and suppress distractors from \(\{\mathcal{L}_m\}_{m=2}^M\). As a result, an instruction-guided attention mechanism emerges, strengthening the cross-modal association and improving localization accuracy in blended scenes.

Overall, the integrated strategy enhances robustness by teaching the model to disregard instruction-irrelevant regions, reinforces text-action associations, and improves multi-target handling. Crucially, \textbf{it establishes invariant relationships between language instructions and action goals by introducing controlled variations in visual scenes}. This is achieved without extra modules or inference overhead, leading to significant gains in generalization, interference resistance, instruction follow-up accuracy, and data efficiency.

\begin{table*}[t]
\centering
\resizebox{\textwidth}{!}{
\begin{tabular}{c c|c|c|c|c|c|c|c}
\hline
\rowcolor[HTML]{D8E4FC} 
\multicolumn{1}{c}{\cellcolor[HTML]{D8E4FC}} & \multicolumn{2}{c}{\cellcolor[HTML]{D8E4FC}} & \multicolumn{6}{c}{\cellcolor[HTML]{D8E4FC}} \\
\rowcolor[HTML]{D8E4FC} 
\multirow{-2}{*}{\cellcolor[HTML]{D8E4FC}} & \multicolumn{2}{c}{\multirow{-2}{*}{\cellcolor[HTML]{D8E4FC}\textbf{Overall}}} & \multicolumn{6}{c}{\multirow{-2}{*}{\cellcolor[HTML]{D8E4FC}\textbf{Success Rate (\%)}}} \\
\cmidrule(lr){2-3} \cmidrule(l){4-9}
\rowcolor[HTML]{D8E4FC} 
\cellcolor[HTML]{D8E4FC}\textbf{Models} & \cellcolor[HTML]{D8E4FC}\textbf{Avg. SR (\%)} $\uparrow$ & \cellcolor[HTML]{D8E4FC}\textbf{Avg. Rank} $\downarrow$ & \cellcolor[HTML]{D8E4FC}All Perturbations & \cellcolor[HTML]{D8E4FC}MO-COLOR & \cellcolor[HTML]{D8E4FC}RO-COLOR & \cellcolor[HTML]{D8E4FC}MO-TEXTURE & \cellcolor[HTML]{D8E4FC}RO-TEXTURE & \cellcolor[HTML]{D8E4FC}MO-SIZE \\
\midrule
PerAct\cite{shridhar2023perceiver}             & 27.9 & 5.71 & 7.2                      & 24.0                    & 29.2                     & 28.8                      & 17.71                      & 35.6                     \\
\rowcolor[HTML]{E7E6E6} 
\cellcolor[HTML]{E7E6E6}RVT\cite{goyal2023rvt}          & \cellcolor[HTML]{E7E6E6}35.4 & \cellcolor[HTML]{E7E6E6}5.28 & \cellcolor[HTML]{E7E6E6}6.4 & \cellcolor[HTML]{E7E6E6}26.0 & \cellcolor[HTML]{E7E6E6}31.3 & \cellcolor[HTML]{E7E6E6}44.8 & \cellcolor[HTML]{E7E6E6}41.1 & \cellcolor[HTML]{E7E6E6}35.3 \\
RVT-2\cite{goyal2024rvt}                & 56.7 & 3.92 & 15.6 $\pm$ 0.8 & 53.0 $\pm$ 0.9 & 54.6 $\pm$ 0.6 & 59.7 $\pm$ 0.7 & 56.7 $\pm$ 1.4 & 60.9 $\pm$ 0.9 \\
\rowcolor[HTML]{E7E6E6} 
\cellcolor[HTML]{E7E6E6}SAM2ACT\cite{fang2025sam2act}               & \cellcolor[HTML]{E7E6E6}61.9 & \cellcolor[HTML]{E7E6E6}2.93 & \cellcolor[HTML]{E7E6E6}18.5 $\pm$ 3.5 & \cellcolor[HTML]{E7E6E6}62.1 $\pm$ 0.9 & \cellcolor[HTML]{E7E6E6}60.9 $\pm$ 0.3 & \cellcolor[HTML]{E7E6E6}58.6 $\pm$ 0.6 & \cellcolor[HTML]{E7E6E6}63.4 $\pm$ 2.1 & \cellcolor[HTML]{E7E6E6}61.8 $\pm$ 0.5 \\
BridgeVLA\cite{li2025bridgevla}                       & 64.0 & 2.14 & 18.7 $\pm$ 2.2           & 60.5 $\pm$ 1.1          & \textbf{63.8 $\pm$ 0.1}           & 63.5 $\pm$ 1.5            & 68.4 $\pm$ 3.3             & 69.3 $\pm$ 1.0           \\
\rowcolor[HTML]{E7E6E6} 
\cellcolor[HTML]{E7E6E6}\textbf{TGM-VLA} & \cellcolor[HTML]{E7E6E6}\textbf{68.8} & \cellcolor[HTML]{E7E6E6}\textbf{1.14} & \cellcolor[HTML]{E7E6E6}\textbf{24.3 $\pm$ 1.1} & \cellcolor[HTML]{E7E6E6}\textbf{65.3 $\pm$ 0.6} & \cellcolor[HTML]{E7E6E6}62.1 $\pm$ 1.0 & \cellcolor[HTML]{E7E6E6}\textbf{72.1 $\pm$ 0.6} & \cellcolor[HTML]{E7E6E6}\textbf{72.4 $\pm$ 1.0} & \cellcolor[HTML]{E7E6E6}\textbf{70.4 $\pm$ 0.9} \\
\midrule
\rowcolor[HTML]{D8E4FC} 
\cellcolor[HTML]{D8E4FC}\textbf{Models} & \cellcolor[HTML]{D8E4FC}RO-SIZE & \cellcolor[HTML]{D8E4FC}Light Color & \cellcolor[HTML]{D8E4FC}Table Color & \cellcolor[HTML]{D8E4FC}Table Texture & \cellcolor[HTML]{D8E4FC}Distractor & \cellcolor[HTML]{D8E4FC}Background Texture & \cellcolor[HTML]{D8E4FC}RLBench & \cellcolor[HTML]{D8E4FC}Camera Pose \\
\midrule
PerAct\cite{shridhar2023perceiver}             & 29.3 & 29.1 & 30.4 & 23.2 & 27.1 & 33.5 & 39.4 & 36.3 \\
\rowcolor[HTML]{E7E6E6} 
\cellcolor[HTML]{E7E6E6}RVT\cite{goyal2023rvt}          & \cellcolor[HTML]{E7E6E6}40.5 & \cellcolor[HTML]{E7E6E6}34.0 & \cellcolor[HTML]{E7E6E6}30.0 & \cellcolor[HTML]{E7E6E6}45.2 & \cellcolor[HTML]{E7E6E6}18.8 & \cellcolor[HTML]{E7E6E6}46.4 & \cellcolor[HTML]{E7E6E6}53.4 & \cellcolor[HTML]{E7E6E6}42.2 \\
RVT-2\cite{goyal2024rvt}                & 53.4 $\pm$ 1.5 & 58.0 $\pm$ 1.1 & 62.6 $\pm$ 0.9 & 56.6 $\pm$ 0.9 & 60.8 $\pm$ 0.5 & 68.7 $\pm$ 1.1 & 68.8 $\pm$ 1.3 & 64.4 $\pm$ 0.5 \\
\rowcolor[HTML]{E7E6E6} 
\cellcolor[HTML]{E7E6E6}SAM2ACT\cite{fang2025sam2act}               & \cellcolor[HTML]{E7E6E6}59.0 $\pm$ 2.4 & \cellcolor[HTML]{E7E6E6}70.3 $\pm$ 0.9 & \cellcolor[HTML]{E7E6E6}71.8 $\pm$ 0.3 & \cellcolor[HTML]{E7E6E6}68.2 $\pm$ 2.3 & \cellcolor[HTML]{E7E6E6}62.3 $\pm$ 0.7 & \cellcolor[HTML]{E7E6E6}68.6 $\pm$ 0.7 & \cellcolor[HTML]{E7E6E6}70.8 $\pm$ 1.4 & \cellcolor[HTML]{E7E6E6}69.7 $\pm$ 0.8 \\
BridgeVLA\cite{li2025bridgevla}                       & 61.7 $\pm$ 0.8 & 69.7 $\pm$ 1.2 & \textbf{75.7 $\pm$ 0.9} & 71.3 $\pm$ 0.7 & 51.8 $\pm$ 1.5 & 74.8 $\pm$ 1.0 & 73.1 $\pm$ 0.2 & 73.8 $\pm$ 0.3 \\
\rowcolor[HTML]{E7E6E6} 
\cellcolor[HTML]{E7E6E6}\textbf{TGM-VLA} & \cellcolor[HTML]{E7E6E6}\textbf{69.4 $\pm$ 2.1} & \cellcolor[HTML]{E7E6E6}\textbf{71.4 $\pm$ 0.7} & \cellcolor[HTML]{E7E6E6}73.6 $\pm$ 0.8 & \cellcolor[HTML]{E7E6E6}\textbf{71.5 $\pm$ 0.6} & \cellcolor[HTML]{E7E6E6}\textbf{74.3 $\pm$ 0.8} & \cellcolor[HTML]{E7E6E6}\textbf{78.9 $\pm$ 1.2} & \cellcolor[HTML]{E7E6E6}\textbf{79.6 $\pm$ 0.7} & \cellcolor[HTML]{E7E6E6}\textbf{78.7 $\pm$ 0.9} \\
\bottomrule
\end{tabular}
}
\vspace{3pt}
\caption{\textbf{Results on the COLOSSEUM Benchmark.} This table shows the success rates across 14 generalization settings (12 unseen environmental perturbations + all perturbations + RLBench). 
The “Avg. Rank” column reports the average rank of each method across all perturbations, where lower values indicate better overall performance.}
\label{tab:colosseum}
\vspace{-5pt}
\end{table*}

\section{EXPERIMENTS}

\subsection{Settings}
We conduct experiments on RLBench \cite{james2020rlbench}, with tasks simulated in CoppeliaSim \cite{rohmer2013v} using a Franka Panda robot. The observations are four calibrated RGB-D views from the front, left shoulder, right shoulder, and wrist. Consistent with prior work \cite{shridhar2023perceiver,gervet2023act3d,goyal2023rvt,3d-da,goyal2024rvt}, we evaluate performance on 18 tasks covering three types: non-prehensile manipulation, pick-and-place, and high-precision insertion. Each task provides 100 expert demonstrations, annotated with language instructions and keyframes. Models are assessed by their task success rate over 25 trials, with a maximum of 25 actions per trial. To ensure statistical robustness, we repeat this evaluation four times and report the mean success rate along with the standard deviation across runs.

To further evaluate the generalization capabilities of TGM-VLA, we perform experiments on COLOSSEUM benchmark \cite{pumacay2024colosseum}. Our model is trained on the standard RLBench but tested in environments with 12 unseen perturbations, including changes to object texture, color, size, background, lighting, distractors, and camera pose. In total, COLOSSEUM provides 20,371 unique perturbed task instances to thoroughly examine model generalization. Our procedure includes: 1) training on 20 RLBench tasks (100 demonstrations per task), 2) testing each task over 25 trials per perturbation type, and 3) averaging the success rate across all tasks for each perturbation. We also report results on the original RLBench tasks and a challenging scenario combining all 12 perturbations.

\subsection{Main Results}
Table \ref{tab:rlbench_tab} compares TGM-VLA with prior keyframe-based 3D VLA methods on the RLBench benchmark. Overall, TGM-VLA achieves a new state-of-the-art \textbf{average success rate of 90.5\%}, significantly surpassing the previous best-performing method BridgeVLA (\textbf{88.2\%}). Notably, our method attains this performance using only \textbf{2$\times$A6000 for one day} of hardware resources, which corresponds to a \textbf{5$\times$ training speed improvement} over the baseline SAM2ACT, which requires 2$\times$A6000 for five days. A closer inspection of individual tasks reveals that TGM-VLA achieves the highest success rate in \textbf{11 out of 18} tasks. The most substantial improvements are observed in precision-demanding and long-horizon tasks: on \textit{Stack Cups}, TGM-VLA reaches 87.0\% compared to BridgeVLA's 81.6\% (+5.4\%); on \textit{Sort Shape}, it attains 80.0\% versus BridgeVLA's 60.8\% (+19.2\%). These results demonstrate that TGM-VLA achieves SOTA performance while exhibiting superior data utilization efficiency and training speed, significantly accelerating the iteration of subsequent algorithms.

Table \ref{tab:colosseum} evaluates TGM-VLA and prior methods on the COLOSSEUM benchmark, which tests generalization under 12 unseen environmental perturbations. TGM-VLA achieves a state-of-the-art \textbf{average success rate of 68.8\%} and the best \textbf{average rank of 1.14}, surpassing BridgeVLA (64.0\%) and demonstrating superior robustness. Notably, TGM-VLA excels in the most challenging \textbf{"All Perturbations"} setting (24.3\%), outperforming BridgeVLA by 5.6\%, and achieves the highest scores in \textbf{12 out of 14} perturbation categories. These results confirm TGM-VLA's strong generalization capability to diverse visual perturbations.

\begin{table}[t]
\centering
% \small
\footnotesize
\renewcommand{\arraystretch}{0.8}
\resizebox{0.99\columnwidth}{!}{  % 缩放到半栏宽度
\setlength{\tabcolsep}{3.0pt}  % ← 添加这一行：减小列间距
\begin{tabular}{@{}ccccc@{}}  % ← 修改这一行：添加@{}去除两侧边距
\toprule
\rowcolor[HTML]{D8E4FC} 
\multirow{2}{*}{\textbf{Method}} & \multicolumn{2}{c}{\textbf{Training}} & \multicolumn{2}{c}{\textbf{Testing}} \\
\cmidrule(lr){2-3} \cmidrule(lr){4-5}
\rowcolor[HTML]{D8E4FC} 
& \textbf{Stack Cups} & \textbf{Setup Chess} & \textbf{Stack Cups} & \textbf{Setup Chess} \\
\midrule
\textbf{w/o CIB} & 80.0$\pm$3.0 & 16.0$\pm$3.8 & 74.0$\pm$4.6 & 12.0$\pm$2.3 \\
\addlinespace
\rowcolor[HTML]{E7E6E6} 
\textbf{Ours} & 94.0$\pm$1.0 & 46.0$\pm$7.3 & 87.0$\pm$2.0 & 30.0$\pm$4.0 \\
\addlinespace
% \rowcolor[HTML]{E7E6E6} 
\textbf{Improvement} & \textbf{+14.0} & \textbf{+30.0} & \textbf{+13.0} & \textbf{+18.0} \\
\bottomrule
\end{tabular}
}
\vspace{1pt}
\caption{\textbf{Ablation Study on Color Inversion.} 
"w/o CIB" denotes ablations without a color inversion branch. 
 }
\label{tab:ablation_inv}
\vspace{-15pt}
\end{table}

\subsection{Ablation Study}

To validate the effectiveness of \textbf{Cross-Task Mixup (CTM)} and \textbf{Intra-Task Mixup (ITM)}, we conduct systematic ablation studies. The full model achieves the best average success rate of 90.5\%. Removing CTM or ITM reduces performance to \textbf{88.2\% (-2.3\%)} and \textbf{88.8\% (-1.7\%)}, respectively, indicating that both strategies contribute positively, with CTM offering a slightly larger gain. CTM excels in complex tasks requiring multi-step planning or precise spatial reasoning (e.g., +7-8\% improvement on Sort Shape and Stack Blocks), as it trains the model to focus on task-relevant features. ITM provides robust generalization across most tasks by learning a broader solution space, making the model less sensitive to minor variations.

We further conduct an ablation study to evaluate the \textbf{color inversion branch (CIB)}, which enhances perception for low-contrast objects. As shown in Table \ref{tab:ablation_inv}, our full model achieves significant improvements across both training and testing environments. The gains are most prominent on \textit{Setup Chess} task, where training performance rises from \textbf{16.0\% to 46.0\% (+30.0\%)} and testing performance from \textbf{12.0\% to 30.0\% (+18.0\%)}. Consistent improvements are also observed on \textit{Stack Cups}, demonstrating the robustness and generalizability of the proposed branch.

\begin{figure}[]
  \centering
  \includegraphics[width=0.99\columnwidth]{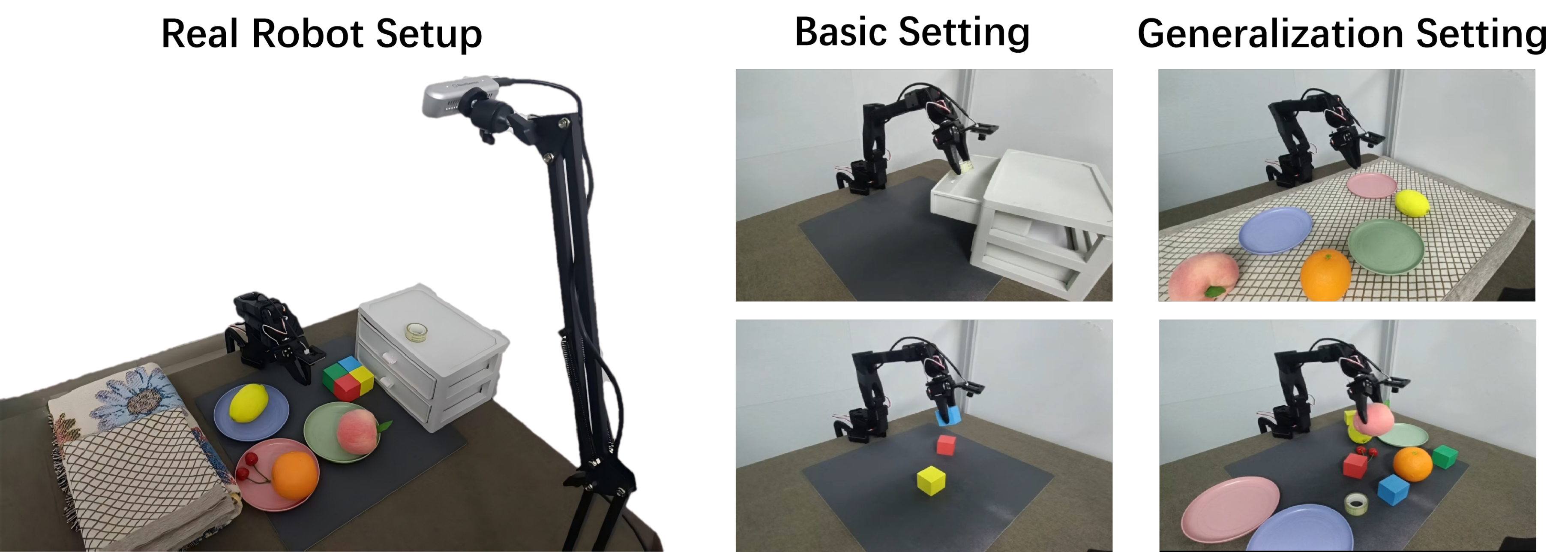}
    \caption{ Real Robot Experimental Setup.
    }
  \label{fig:real}
  \vspace{-5pt}
\end{figure}

\begin{table}[t]
\centering
\footnotesize
\renewcommand{\arraystretch}{0.8}
\resizebox{1.0\columnwidth}{!}{  % 缩放到半栏宽度
\setlength{\tabcolsep}{2.2pt}
\begin{tabular}{@{}lccccc@{}}
\toprule
\rowcolor[HTML]{D8E4FC} 
\multirow{2}{*}{\textbf{Method}} & \multicolumn{3}{c}{\textbf{Basic Tasks}} & \multicolumn{2}{c}{\textbf{Generalization}} \\
\cmidrule(lr){2-4} \cmidrule(lr){5-6}
\rowcolor[HTML]{D8E4FC} 
& {\scriptsize \textbf{Stack Blocks}} & {\scriptsize \textbf{Put in Drawer}} & {\scriptsize \textbf{Put on Plate}} & {\scriptsize \textbf{Distractor}} & {\scriptsize \textbf{Background}} \\
\midrule
\textbf{w/o TGM} & 60\% & 60\% & 80\% & 40\% & 60\% \\
\addlinespace
\rowcolor[HTML]{E7E6E6} 
\textbf{TGM-VLA} & 80\% & 70\% & 90\% & 90\% & 80\% \\
\addlinespace
{\scriptsize \textbf{Improvement}} & \textbf{+20\%} & \textbf{+10\%} & \textbf{+10\%} & \textbf{+50\%} & \textbf{+20\%} \\
\bottomrule
\end{tabular}
}  % resizebox结束
% \vspace{3pt}
\caption{Results in the real world. "w/o TGM" denotes ablations without task-guided mixup technique.}
\label{tab:ablation_real}
\vspace{-15pt}
\end{table}

\subsection{Real World Experiments}
We also validate our model on a real-world setup using an SO101 robot arm with a static D435 RGB-D camera, as illustrated in Fig.~\ref{fig:real}. The experiments cover five manipulation tasks that range from basic operations to more challenging scenarios involving visual perturbations. For each task, we define 5–8 keyframes and collect 10 expert trajectories via teleoperation for training. During evaluation, the robot sequentially reaches the predicted key poses; we report success rates averaged over 10 trials per task.

Table~\ref{tab:ablation_real} compares the performance of TGM-VLA with a baseline that omits task-guided mixup (w/o TGM). Our full model consistently outperforms the baseline across all five tasks. On basic manipulation tasks (Stack Blocks, Put in Drawer, Put on Plate), TGM-VLA achieves success rates of 80\%, 70\%, and 90\%, respectively, surpassing the baseline by +20\%, +10\%, and +10\%. The gains are even more pronounced on generalization tasks: under distractor perturbation, the success rate increases from 40\% to 90\% (+50\%), and under background changes, it rises from 60\% to 80\% (+20\%). These substantial improvements demonstrate that task-guided mixup is critical for robustness to environmental perturbations, as it encourages the model to focus on task-relevant features while ignoring task-irrelevant variations.

\section{CONCLUSION}

This paper presents TGM-VLA, a sampling-efficient and robust 3D vision-language-action model that achieves SOTA performance by addressing three key bottlenecks: an optimized sampling strategy, a color inversion branch, and a task-guided mixup technique. These innovations directly tackle the core challenges of data redundancy, perceptual ambiguity, and instruction underutilization that hinder existing methods. Looking forward, while our handcrafted heuristic sampling proves effective, it also invites future research into adaptive sampling and physics-aware data augmentation to further bridge the sim-to-real gap. This work underscores the substantial impact of data-centric innovations in developing more capable and generalizable robotic agents.

\addtolength{\textheight}{-12cm}   % This command serves to balance the column lengths

%%%%%%%%%%%%%%%%%%%%%%%%%%%%%%%%%%%%%%%%%%%%%%%%%%%%%%%%%%%%%%%%%%%%%%%%%%%%%%%%

% \section*{ACKNOWLEDGMENT}
% This work was supported by Zhongbing Intelligent Innovation Research Institute.

\bibliographystyle{IEEEtran}
\bibliography{root}

\end{document}